\documentclass[10pt,journal,compsoc]{IEEEtran}
\usepackage[hyphens]{url}  % DO NOT CHANGE THIS
\usepackage{graphicx} % DO NOT CHANGE THIS

\usepackage{amsfonts} 
\usepackage{subfigure}
\usepackage{multirow}

\usepackage{enumitem}
\usepackage{amsmath}
\usepackage{amsthm}

\newcommand{\ie}{\emph{i.e., }}
\newcommand{\eg}{\emph{e.g., }}

\newcommand{\wrt}{\emph{w.r.t. }}

\ifCLASSOPTIONcompsoc
  \usepackage[nocompress]{cite}
\else
  \usepackage{cite}
\fi
\ifCLASSINFOpdf
\else
\fi
\begin{document}
\title{CatGCN: Graph Convolutional Networks with Categorical Node Features}

\author{Weijian Chen, Fuli Feng, Qifan Wang, Xiangnan He, Chonggang Song, Guohui Ling, Yongdong Zhang
\IEEEcompsocitemizethanks{
	\IEEEcompsocthanksitem Weijian Chen, Xiangnan He and Yongdong Zhang are with the School of Information Science and Technology, University of Science and Technology of China, Hefei, China. E-mail: naure@mail.ustc.edu.cn, xiangnanhe@gmail.com, zhyd73@ustc.edu.cn. \protect\\
	\IEEEcompsocthanksitem Fuli Feng is with the School of Computing, National University of Singapore, Computing 1, Computing Drive, 117417, Singapore. E-mail: fulifeng93@gmail.com. \protect\\
	\IEEEcompsocthanksitem Qifan Wang is with Google Research. E-mail: wqfcr@google.com. \protect\\
	\IEEEcompsocthanksitem Chonggang Song, Guohui Ling are with Tencent WeChat. E-mail: jerrycgsong@tencent.com, randyling@tencent.com. \protect\\
	\IEEEcompsocthanksitem Fuli Feng and Yongdong Zhang are corresponding authors. \protect\\

}}

\markboth{Journal of \LaTeX\ Class Files,~Vol.~14, No.~8, August~2015}%
{Shell \MakeLowercase{\textit{et al.}}: Bare Demo of IEEEtran.cls for Computer Society Journals}



\IEEEtitleabstractindextext{%
\begin{abstract}
Recent studies on Graph Convolutional Networks (GCNs) reveal that the initial node representations (i.e., the node representations before the first-time graph convolution) largely affect the final model performance.
However, when learning the initial representation for a node, most existing work linearly combines the embeddings of node features, without considering the interactions among the features (or feature embeddings).
We argue that when the node features are categorical, e.g., in many real-world applications like user profiling and recommender system, feature interactions usually carry important signals for predictive analytics. 
Ignoring them will result in suboptimal initial node representation and thus weaken the effectiveness of the follow-up graph convolution. 
In this paper, we propose a new GCN model named CatGCN, which is tailored for graph learning on categorical node features. Specifically, we integrate two ways of explicit interaction modeling into the learning of initial node representation, i.e., local interaction modeling on each pair of node features and global interaction modeling on an artificial feature graph. 
We then refine the enhanced initial node representations with the neighborhood aggregation-based graph convolution.
We train CatGCN in an end-to-end fashion and demonstrate it on the task of node classification. 
Extensive experiments on three tasks of user profiling (the prediction of user age, city, and purchase level) from Tencent and Alibaba datasets validate the effectiveness of CatGCN, especially the positive effect of performing feature interaction modeling before graph convolution. 
\end{abstract}

}

\maketitle

\IEEEdisplaynontitleabstractindextext

% \fi
\IEEEpeerreviewmaketitle

\IEEEraisesectionheading{\section{Introduction}
\label{sec:introduction}}

\IEEEPARstart{G}{CNs} have become a promising technique in various applications~\cite{zhang2020deep}, such as recommender system~\cite{PinSAGE,LightGCN,liu2020modelling}, user profiling~\cite{rahimi2018semi,HGAT} and text mining~\cite{yao2019graph}.
The main idea of graph convolution is to relate the representations of nodes based on the graph structure s.t. connected nodes should have similar representations, which can be seen as enforcing the smoothness constraint in the representation space. 
For example, the standard GCN~\cite{GCN} performs layer-wise representation relating as:
\begin{equation}
    \textbf{H}^{(l+1)} = \sigma ( \widetilde{\textbf{A}} \textbf{H}^{(l)} \textbf{W}^{(l)} ),
\end{equation}
where $\textbf{H}^{(l)}$ is the node representation matrix of the $l$-th layer, $\widetilde{\textbf{A}}$ is the normalized graph adjacency matrix, and $\textbf{W}^{(l)}$ is the weight matrix of the $l$-th layer (i.e., trainable model parameters of GCN). 
The $\textbf{H}^{(0)}$ matrix stores the input features of nodes, e.g., the frequency of words of a document node~\cite{GCN}. 
We term $\textbf{H}^{(0)} \textbf{W}^{(0)}$ as the  \textit{initial node representation}, which performs linear transformation on the input features of each node and obtains a representation for the follow-up graph convolution operation. 

Assuming the input node features are categorical, the feature matrix $\textbf{H}^{(0)}$ is then high-dimensional yet sparse, in which each non-zero entry denotes the categorical feature value of a node. We can then understand the initial representation of a node (i.e., a row vector of $\textbf{H}^{(0)} \textbf{W}^{(0)}$) as linearly combining the embedding vectors of the node's categorical features (i.e., the row vectors of $\textbf{W}^{(0)}$). With such a linear combination, the interactions among feature embeddings are not considered. Although the weight matrices of the following layers (e.g., $\textbf{W}^{(1)}$ and $\textbf{W}^{(2)}$) may capture some interactions, the process is rather implicit and ineffective for learning cross feature effects~\cite{NFM,CrossNet}. 

The tying of feature transformation and neighborhood aggregation in each graph convolution layer also limits the representation quality. Decoupled GCN such as APPNP first uses a conventional neural network on node features to obtain a representation vector (the same size as the label space) for each node; it then performs \textit{pure neighborhood aggregation} --- with no weight matrices and other trainable parameters --- to refine the representation vector for prediction. Their strong performance inspires us to believe that the better the initial node representation is, the more benefits the follow-up graph convolution (or equivalently, neighborhood aggregation) can achieve. 
This is because that, the benefits brought by neighborhood aggregation and feature transformation are orthogonal --- one exploits the signal from a node's neighbors whereas the other mostly depends on the features of a node itself. 
As such, if better (e.g., more discriminative) representation for a node can be obtained by leveraging its input features, the performance after neighborhood aggregation should be better.

Although much effort has been devoted to inventing new GCN models, they mostly focus on graph convolution operations~\cite{APPNP,SGC}. 
To our knowledge, seldom research has considered improving the ability of GCN from the perspective of initial node representation, especially for categorical node features. 
In fact, many real-world applications have categorical features as raw data more commonly than continuous features, which are mostly restricted to multimedia content like images and videos.
For example, in recommender systems, nodes are users and items that are normally described by user demographics (age, gender, interest tags) and item profiles (category, brand, etc.) \cite{wu2021survey}; 
in search engines, nodes are queries and documents that are described by bag-of-words or n-grams;
on rating sites, content nodes such as music and videos are described with tags.
For such categorical features, the interactions among features --- e.g., the co-occurrence of multiple features --- could contain important signal on the node's properties~\cite{NFM,CrossNet,WideDeep}. 
However, most GCNs apply a simple sum of feature embeddings as the initial node representation, which we believe is insufficient to model feature interactions and results in suboptimal node representation. 

In this work, we explore how improved representation learning from categorical node features benefits GCN. We propose a new model named CatGCN, which integrates two kinds of explicit feature interactions into initial node representation learning: 1) local multiplication-based interaction on each pair of node features, and 2) global addition-based interaction on an artificial feature graph. 
We prove that in the artificial feature graph, performing one graph convolution layer with tunable self-connections can capture the interactions among all features. 
We then feed the enhanced initial node representations into a simplified/light GCN~\cite{SGC,LightGCN} that performs neighborhood aggregation only to exploit the graph structure for node representation learning. The CatGCN is end-to-end trainable, such that all parameters in the initial node representation learning, and follow-up graph convolution and prediction layers can be optimized towards the final task. 

The main contributions of the paper are summarized as follows:
\begin{itemize}
 \item We emphasize the importance of tailoring GCNs for categorical node features, especially by modeling the interactions among features before graph convolution. 
 \item We propose CatGCN, which performs two kinds of feature interaction modeling to enhance the initial node representations. 
 \item We conduct experiments on user profiling tasks on large-scale datasets, verifying the positive effect of performing feature interaction modeling before graph convolution. 
\end{itemize}

\section{Methodology}
\label{sec:methodology}

\begin{table}[]
	\caption{Terms and notations used in the paper.}
	\vspace{-0.3cm}
	\label{tab:terms}
	\resizebox{0.48\textwidth}{!}{
		\begin{tabular}{c|l}
			\hline
			Symbol  & Definition \\ \hline \hline
			$\mathbf{A} \in \mathbb{R}^{N \times N}$ & the adjacency matrix of a graph \\
			$\mathbf{D} \in \mathbb{R}^{N \times N}$ & the degree matrix of a graph \\
			$\mathbf{I} \in \mathbb{R}^{N \times N}$ & the identity matrix of a graph  \\
			$\mathbf{H} \in \mathbb{R}^{N \times C}$ &  the  initial  node  representations \\
			$\mathbf{Y} \in \mathbb{R}^{N \times C}$ & the final  node  representations \\
			$\mathbf{x}_u \in \mathbb{R}^{d}$ & the categorical features of node $u$ \\
			$\mathbb{S}_u = \{i | x_i^u \neq 0\}$ & the set of nonzero features of node $u$ \\
			$\mathbf{E} \in \mathbb{R}^{|\mathbb{S}| \times D}$ & the embedding matrix of categorical features \\
			$\mathbf{e} \in \mathbb{R}^{D}$ & the embedding vector of one categorical feature \\
			$\mathbf{P} \in \mathbb{R}^{|\mathbb{S}| \times |\mathbb{S}|}$ & the adjacency matrix of one node's artificial feature graph  \\
			$\mathbf{Q}  \in \mathbb{R}^{|\mathbb{S}| \times |\mathbb{S}|}$ & the degree matrix  of one node's artificial feature graph  \\
			$\mathbf{O}  \in \mathbb{R}^{|\mathbb{S}| \times |\mathbb{S}|}$ & the identity matrix  of one node's artificial feature graph  \\
			$\mathbf{W}, \mathbf{W}_l, \mathbf{W}_g$ &  the  weight matrices \\
			$\mathbf{b}, \mathbf{b}_l, \mathbf{b}_g$ & the bias vectors  \\
			$\mathcal{\mathbf{\Theta}}$ & all model parameters\\
			$\odot$ & element-wise product \\
			$\sigma$ & a non-linear activation function \\
			\hline
		\end{tabular}
	}
\end{table}

We describe our method under the setting of node classification~\cite{GCN}, whereas the idea is generally applicable to GCNs for other tasks like link prediction~\cite{LightGCN,zhang2018link} and community detection~\cite{chen2019supervised}.
The graph structure is represented as an adjacency matrix $\mathbf{A} \in \mathbb{R}^{N \times N}$ where $N$ is the number of nodes. The main consideration of our work is that, each node $u$ in the graph is described by \textbf{categorical features} $\mathbf{x}_u \in \mathbb{R}^{d}$ ($d$ is the number of total features) where an entry $x_i^u = 0$ means the $i$-th feature value does not exist in the node (e.g., a female user cannot have ``male'' in her feature values). 
For a categorical feature vector $\mathbf{x}_u$, we denote the set of nonzero features as $\mathbb{S}_u = \{i | x_i^u \neq 0\}$.
We summarize the symbols used in the paper in Table \ref{tab:terms}.

\subsection{Overall framework} 
\begin{figure*}
 \centering
 \subfigure{
 \includegraphics[width=1.0\textwidth]{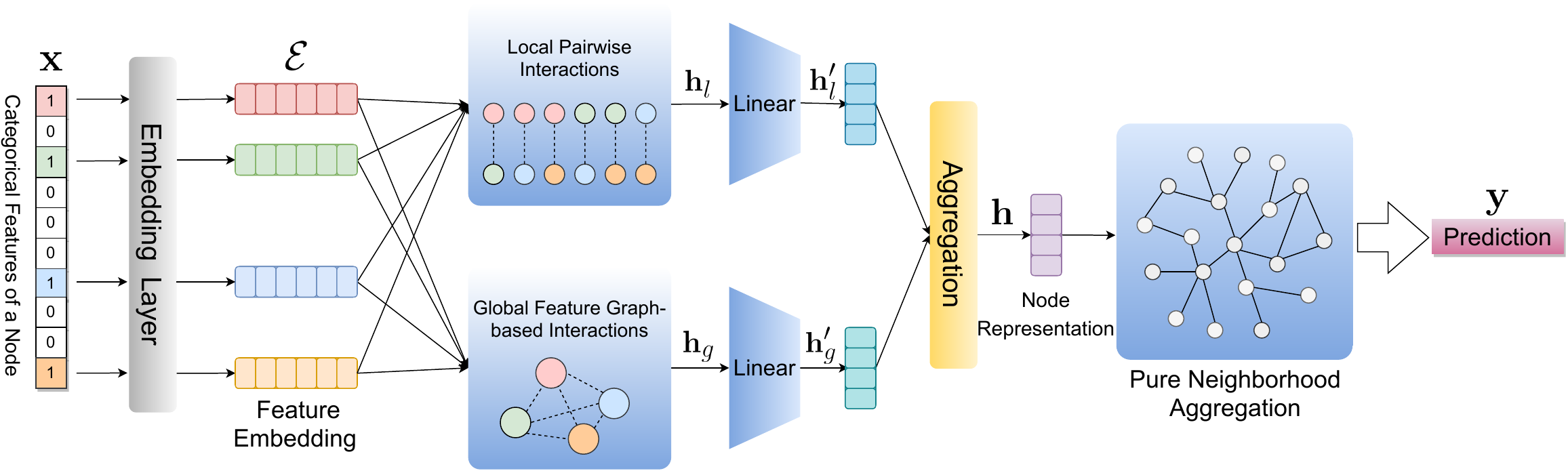}}
 \caption{The framework of CatGCN where an example node is taken to demonstrate the procedure of computing the initial node representation.
 Here, the feature number ($d$), embedding size ($D$), and prediction classes ($C$) are set as 10, 6, and 4, respectively.
 }
 \label{fig:framework}
\end{figure*}

The target of CatGCN is to improve initial node representations by explicitly incorporating the interactions of categorical features in a lightweight and efficient manner. 
As illustrated in Figure~\ref{fig:framework}, CatGCN separately exploits the categorical features of a node itself and the signal from its neighbors. 
In particular, CatGCN first learns initial node representation $\mathbf{h}_u$ from its categorical features $\mathbf{x}_u$ with dedicated interaction modeling (detailed in Section~\ref{sec:ft}). 
CatGCN then performs \textit{pure neighborhood aggregation} (PNA) over the graph structure, which is formulated as:
\begin{equation}
\mathbf{Y}={PNA}(\mathbf{\widetilde{A}}, \mathbf{H}, L), ~~~\mathbf{\widetilde{A}}=\mathbf{{D}}^{-\frac{1}{2}}\mathbf{\hat{A}}\mathbf{{D}}^{-\frac{1}{2}}, ~~~\mathbf{\hat{A}}=\mathbf{A}+\mathbf{I},
\end{equation}
where $\mathbf{H}$ and $\mathbf{Y} \in \mathbb{R}^{N \times C}$ denote the initial node representations and final node representations with $L$-hop neighbors aggregated.
Here, $C$ is the number of prediction classes, $\mathbf{\hat{A}}$ is the adjacency matrix $\mathbf{A}$ with self-loops added~(corresponding to the identity matrix $\mathbf{I}$).
$\mathbf{\hat{A}}$ is normalized by node degrees which are organized into a diagonal degree matrix $\mathbf{{D}}$.
Along the development of graph convolution operations, the $L$-hop neighborhood aggregation is either implemented in an iterative manner with $L$ repeats of $\mathbf{\widetilde{A}}\mathbf{H}^{(l - 1)}$ ($\mathbf{H}^{0} = \mathbf{H}$)~\cite{GCN}, or implemented in a simplified manner $\mathbf{\widetilde{A}}^{L}\mathbf{H}$ where $\mathbf{\widetilde{A}}^{L}$ is calculated as a pre-processing~\cite{SGC}. 
Following the principle of lightweight design, CatGCN adopts the simplified implementation to avoid the memory overhead of storing intermediate variables and the repeated computation during training. 

Similar as standard GCNs, CatGCN is learned in an end-to-end manner by optimizing an objective function:
\begin{equation}
\mathcal{L}=\sum_{u \in \mathbb{U}}l(\mathbf{g}_{u}, \mathbf{\tilde{y}}_u) + \eta \parallel\mathbf{\Theta}\parallel^2_F, ~~~\mathbf{\tilde{y}}_u = softmax(\mathbf{y}_u),
\end{equation}
where $l(\cdot)$ is a classification loss such as cross-entropy~\cite{GCN} over the training set $\mathbb{U}$ of labeled nodes. The final node representation $\mathbf{y}_u$ of node $u$ is normalized to be a distribution over prediction labels $\mathbf{\tilde{y}}_u$. The one-hot vector $\mathbf{g}_{u} \in \mathbb{R}^{C}$ denotes the ground-truth of node $u$.
$\mathbf{\Theta}$ represents all model parameters, and $\eta$ is a hyper-parameter to balance the effect of loss and regularization. 
In the following, we introduce the learning of initial node representation $\mathbf{h}_u$ from its categorical features $\mathbf{x}_u$, the subscript $u$ is thus omitted for the briefness of notations.

\subsection{Interaction modeling of categorical features}
\label{sec:ft}
Inspired by the effectiveness of explicit feature interaction modeling~\cite{FM,NFM,xgboost} in predictive analytics with categorical features, CatGCN focuses on improving the quality of initial node representations $\mathbf{h}$ via feature interaction modeling. 
To thoroughly capture feature interactions, our first belief is that separately modeling the feature interactions of different forms is essential since they convey different signals. 
Here, we consider the feature interactions of two forms: 1) local interaction between features, i.e., within a feature pair; and 2) global interaction amongst the whole feature set $\mathbb{S}$.
Although much effort has been devoted to modeling local feature interactions~\cite{FM,AFM,NFM}, seldom research has considered the modeling of global interactions. 

To bridge this gap, CatGCN integrates both local and global interactions into initial node representation learning.
Specifically, as shown in Figure~\ref{fig:framework}, CatGCN consists three main modules:
\begin{itemize}[leftmargin=*]
    \item \textbf{Feature embedding.} CatGCN first projects the categorical features into feature embeddings, \ie $\mathbf{x} \longrightarrow \mathcal{E} = \left\{\mathbf{e}_i | i \in \mathbb{S} \right\}$\footnote{Note that each categorical feature in $\mathbb{S}$ is associated with the same node, and their weight can be regarded as 1, so the embedding weight in the following formulas is omitted. In cases where features come with non-binary weights, they can be used to multiply the corresponding embeddings as a preprocessing operation here.}, so as to capture the relative relations among features in the embedding space. Note that $\mathbf{e}_i \in \mathbb{R}^{D}$ denotes the embedding of categorical feature $i$. 
    \item \textbf{Interaction modeling.} Upon the feature embeddings, CatGCN explicitly models the local interaction and global interaction with multiplication-based operation and addition-based operation, respectively. 
    \item \textbf{Fusion.} Lastly, CatGCN is equipped with a fusion module to unify the benefits from both local and global interactions.
\end{itemize}

\subsubsection{Local interaction modeling}
For local feature interaction modeling, effective feature combinations can be mined to enrich input information.
For example, people with pair-wise feature \textit{gender\_age=\{male, 20-25\}} are more likely to be digital enthusiasts. 
This combination of features is more discriminating than either \textit{gender=\{male\}} or \textit{age=\{20-25\}} alone.
The multiplication operation has been widely used to capture the correlation between entities in various tasks such as machine translation~\cite{vaswani2017attention}, recommendation system~\cite{FM}, and text classification~\cite{Cross-GCN}.
In this way, local feature interactions are typically formulated as the element-wise product of feature embeddings. 
A representative operation is the bilinear interaction pooling~\cite{NFM}, 
\begin{equation}
\mathbf{h}_l = \sum_{i, j \in \mathbb{S}~\&~ j > i} \mathbf{e}_i\odot \mathbf{e}_j = \frac{1}{2} \left[ (\sum_{i \in \mathbb{S}}\mathbf{e}_i)^2 - \sum_{i \in \mathbb{S}} \mathbf{e}_i^2 \right],
\label{Bi-Interaction}
\end{equation}
which aggregates the element-wise product on each pair of (different) feature embeddings.
Here, $\mathbf{e}^2$ denotes $\mathbf{e} \odot \mathbf{e}$, and $\mathbf{h}_l$ denotes the initial node representation learned through local feature interaction modeling.
Directly executing the operation has a quadratic time complexity w.r.t. the feature number (i.e., $O(|\mathbb{S}|^2)$), which can be reduced to linear complexity $O(|\mathbb{S}|)$ with an equivalent reformulation~\cite{FM,NFM}.
This is an appealing property of bi-interaction pooling, which models pairwise interactions but with a linear complexity. 
After this operation,  we can obtain more useful interactive features. 
Specifically, $\mathbb{|S|}$ categorical features can be extended to $\mathbb{|S|}$ ($\mathbb{|S|}-1)/2$, which enriches the available information and is obviously of great value to the sparse features.
Note that one can also perform high-order interaction modeling in a similar way~\cite{HOFM}, but the complexity increases polynomially and might be numerically unstable, so we do not further explore it here.  

\subsubsection{Global interaction modeling}
Distinct from local interaction modeling, the purpose of global interaction modeling is to capture the node peculiarity information related to the predicted target.
In real scenarios, the categorical features associated with a node are often diverse, potentially reflecting the different peculiarities of the node.
For instance, a user's purchase history includes laptops, cellphones, drones, running shoes, and sportswear, which indicates the peculiarity information of digital products and sports.
Such peculiarities can be closely related to the prediction target of user interest, e.g., digital products indicates the user is a ``digital enthusiast''. 
Therefore, we need to filter out the latent peculiarities from the feature set $\mathbb{S}$ to facilitate the prediction.
Inspired by the theory of spectral analysis~\cite{ortega2018graph}, our belief is that the latent peculiarity lies in a certain frequency in the spectral domain.
Our key consideration for the global interaction modeling is thus to uncover the signal along the spectrum and strengthen the signal closely associated with the prediction target at a particular frequency, to enhance the quality of the initial node representation.
Considering the ability of GCN to filter frequency~\cite{SGC}, we achieve the global interaction modeling with a carefully designed GCN.

We first use an artificial graph $(\mathbf{P}, \mathbf{E})$ to represent the nonzero features of the node $\mathbb{S}$ where $\mathbf{P} \in \mathbb{R}^{|\mathbb{S}| \times |\mathbb{S}|}$ is the adjacency matrix and $\mathbf{E} \in \mathbb{R}^{|\mathbb{S}| \times D}$ includes the embeddings of features in $\mathbb{S}$. 
$\mathbf{E}$ is initialized with Xaiver~\cite{Xavier}, which is learnable.
As all features in $\mathbb{S}$ have inherent connections (e.g., co-occurrence), the artificial graph is thus a complete graph by natural, and the adjacency matrix $\mathbf{P}$ is an all-ones matrix (with self-loops).
Aiming to capture the global interactions, graph convolution is conducted over the artificial graph.
Formally,
\begin{equation}\notag
\mathbf{\widetilde{P}}=\mathbf{Q}^{-\frac{1}{2}}(\mathbf{P}+\rho\mathbf{O})\mathbf{Q}^{-\frac{1}{2}}=\frac{\mathbf{P}+\rho\mathbf{O}}{|\mathbb{S}|+\rho},
\label{global_1}
\end{equation}
\begin{equation}
\mathbf{h}_g = pool(\sigma(\mathbf{\widetilde{P}}\mathbf{E}\mathbf{W})).
\label{global_2}
\end{equation}
$\mathbf{\widetilde{P}} \in \mathbb{R}^{|\mathbb{S}| \times |\mathbb{S}|}$ is the normalized adjacency matrix with probe coefficient $\rho$ (see Section~\ref{sec: sa} for a detailed theoretical analysis), which adjusts the frequency to be strengthened.
$\mathbf{Q} = (|\mathbb{S}|+\rho)\mathbf{O}  \in \mathbb{R}^{|\mathbb{S}| \times |\mathbb{S}|}$ is the degree matrix of artificial graph where $\mathbf{O} \in \mathbb{R}^{|\mathbb{S}| \times |\mathbb{S}|}$ is an identity matrix.
$\mathbf{h}_g$ denotes the initial node representation learned through global feature interaction modeling.
$\mathbf{W}$ is the weight matrix; $\sigma(\cdot)$ is an activation function such as ReLU; $pool (\cdot)$ is a pooling function such as mean pooling to aggregate the global interactions across features. It should be noted that we model global interactions with only one graph convolution layer, which can largely reduce the memory and computation cost. 
This is because one graph convolution layer can achieve the equivalent effect of multiple layers.

\textbf{Theorem.}
\textit{On graph $\mathbf{\widetilde{P}}$, $K$-hop neighborhood aggregation equals to a $1$-hop aggregation with smaller $\rho$. Formally,}
\begin{equation}
(\frac{\mathbf{P}+\rho_1\mathbf{O}}{|\mathbb{S}|+\rho_1})^K=
\frac{\mathbf{P}+\rho_2\mathbf{O}}{|\mathbb{S}|+\rho_2},\rho_1\geq \rho_2\geq 0,
\label{theorem_a}
\end{equation}
\begin{equation}
\rho_2=
\frac{\rho_1^K}{ \sum_{i=0}^{K-1}C_K^i \rho_1^i |\mathbb{S}|^{K-1-i}}.
\label{theorem_b}
\end{equation}

\paragraph{Proof of the theorem}
\label{sec: prof}
This theorem can be proved by using mathematical induction twice.
We prove the upper half~(i.e., Formula~(\ref{theorem_a})) of this theorem, and the proof of the lower part~(i.e., Formula~(\ref{theorem_b})) is based on the first one.

1) Proof of Formula~(\ref{theorem_a}): 
The normalized adjacency matrix $\mathbf{\widetilde{P}}$ of the categorical feature artificial graph is a symmetric matrix, whose elements $(\{\widetilde{P}_{ij}| 1 \leq i,j \leq |\mathbb{S}|\})$ have such forms:
\begin{equation}
\widetilde{P}_{ij} =
\begin{cases} 
\dfrac{1+\rho_1}{|\mathbb{S}|+\rho_1}, & i=j, \\ 
\dfrac{1}{|\mathbb{S}|+\rho_1}, & i\neq j.
\end{cases}
\end{equation}
Now we prove that the $K$-power of this matrix satisfies Formula~(\ref{theorem_a}).

$\bullet~K = 2$. The quadratic power of $\mathbf{\widetilde{P}}$, i.e., $\mathbf{\widetilde{P}}^2$, has the entries of:
\begin{equation}
\widetilde{P}^2_{ij} =
\begin{cases} 
(\dfrac{1+\rho_1}{|\mathbb{S}|+\rho_1})^2+
\dfrac{{|\mathbb{S}|-1}}{(|\mathbb{S}|+\rho_1)^2}
\\ =
\dfrac{{\rho_1^2+2\rho_1+|\mathbb{S}|}}{(|\mathbb{S}|+\rho_1)^2}=\dfrac{1+\rho_2}{|\mathbb{S}|+\rho_2}, & i=j, \\ 
2(\dfrac{1+\rho_1}{|\mathbb{S}|+\rho_1})(\dfrac{1}{|\mathbb{S}|+\rho_1})+
\dfrac{{|\mathbb{S}|-2}}{(|\mathbb{S}|+\rho_1)^2}
\\ =
\dfrac{{2\rho_1+|\mathbb{S}|}}{(|\mathbb{S}|+\rho_1)^2}=\dfrac{1}{|\mathbb{S}|+\rho_2}, & i\neq j,
\end{cases}
\end{equation}
where $\rho_2 = \frac{{\rho_1^2}}{|\mathbb{S}|+2\rho_1}$ and $\rho_2 \leq \rho_1$. That is to say, by setting the probe coefficient $\rho$ with a small value $\rho_2$, performing $1$-hop propagation  is equivalent to a $2$-hop propagation with probe coefficient of $\rho_1$. 

$\bullet~K > 2$. We assume that the Formula~(\ref{theorem_a}) is correct for $K=k$, that is, we assume that $\mathbf{\widetilde{P}}^{k}$ has the diagonal elements $(1+\rho_k)/(|\mathbb{S}|+\rho_k)$ and the remaining values $1/(|\mathbb{S}|+\rho_k)$.
Under this induction assumption, we must prove that the formula~\ref{theorem_a} is true for its successor, $K=k+1$.
Based on $\mathbf{\widetilde{P}}^{k+1}=\mathbf{\widetilde{P}}^{k}\mathbf{\widetilde{P}}$ and the above induction assumption, the element $\widetilde{P}^{k+1}_{ij}$ of $(k+1)$-power of $\mathbf{\widetilde{P}}$ equals:
\begin{equation}
\begin{cases} 
(\dfrac{1+\rho_k}{|\mathbb{S}|+\rho_k})(\dfrac{1+\rho_1}{|\mathbb{S}|+\rho_1})+
\dfrac{{|\mathbb{S}|-1}}{(|\mathbb{S}|+\rho_k)(|\mathbb{S}|+\rho_1)}
\\ = 
\dfrac{{\rho_k\rho_1+\rho_k+\rho_1+|\mathbb{S}|}}{(|\mathbb{S}|+\rho_k)(|\mathbb{S}|+\rho_1)}, & i=j, \\ 

(\dfrac{1+\rho_k}{|\mathbb{S}|+\rho_k})(\dfrac{1}{|\mathbb{S}|+\rho_1})+
(\dfrac{1}{|\mathbb{S}|+\rho_k})(\dfrac{1+\rho_1}{|\mathbb{S}|+\rho_1}) \\ +
\dfrac{{|\mathbb{S}|-2}}{(|\mathbb{S}|+\rho_k)(|\mathbb{S}|+\rho_1)}
\\ =
\dfrac{{\rho_k+\rho_1+|\mathbb{S}|}}{(|\mathbb{S}|+\rho_k)(|\mathbb{S}|+\rho_1)}, & i\neq j.
\end{cases}
\label{formula_1}
\end{equation}

Now, we need to prove:
\begin{equation}
\widetilde{P}^{k+1}_{ij} =
\begin{cases} 
\dfrac{1+\rho_{k+1}}{|\mathbb{S}|+\rho_{k+1}}, & i=j, \\ 
\dfrac{1}{|\mathbb{S}|+\rho_{k+1}}, & i\neq j.
\end{cases}
\label{formula_2}
\end{equation}

Combine Formula~(\ref{formula_1}) and~Formula~(\ref{formula_2}), we can calculate the probe coefficient $\rho_{k+1}$ that satisfies the Formula~(\ref{theorem_a}), as follows:
\begin{equation}
\rho_{k+1}=
\frac{{\rho_k\rho_1}}{|\mathbb{S}|+\rho_k+\rho_1}
\leq \rho_k.
\label{conclusion_2}
\end{equation}

We have now fulfilled both conditions of the principle of mathematical induction.
The Formula~(\ref{theorem_a}) is therefore true for every natural number $K$. In other words, performing an $1$-hop propagation over the graph can achieve the same effect as performing a $K$-hop propagation.

2) Proof of Formula~(\ref{theorem_b}): 
We follow the same principle to prove the formula.
First, according to the inferred value of $\rho_2$ above, it's easy to find that the Formula~(\ref{theorem_b}) is true when $K=2$, as follows:
\begin{equation}\notag
\rho_2=
\frac{{\rho_1^2}}{|\mathbb{S}|+2\rho_1}=
\frac{\rho_1^2}{ \sum_{i=0}^{2-1}C_2^i \rho_1^i |\mathbb{S}|^{2-1-i}}.
\end{equation}

Next, we assume that the Formula~(\ref{theorem_b}) is correct for $K=k$, formally:
\begin{equation}
\rho_k=
\frac{\rho_1^k}{ \sum_{i=0}^{k-1}C_k^i \rho_1^i |\mathbb{S}|^{k-1-i}}.
\label{assumption}
\end{equation}

With this assumption, we must show that the rule is true for its successor, $K=k+1$.
Based on the conclusion of Formula~(\ref{conclusion_2}) and Formula~(\ref{assumption}) above, we have
\begin{equation}
\begin{aligned}	
\rho_{k+1} & =
\frac{{\rho_k\rho_1}}{|\mathbb{S}|+\rho_k+\rho_1} \\ & =
\frac{\rho_1^{k+1}}
{ \sum_{i=0}^{k-1}C_k^i \rho_1^{i+1} |\mathbb{S}|^{k-1-i}+
\sum_{i=0}^{k-1}C_k^i \rho_1^{i} |\mathbb{S}|^{k-i}+
\rho_1^{k}
} \notag
\end{aligned}
\end{equation}

Now, we need to prove:
\begin{equation}\notag
\rho_{k+1}=
\frac{\rho_1^{k+1}}{ \sum_{i=0}^{k}C_{k+1}^i \rho_1^i |\mathbb{S}|^{k-i}}
\end{equation}

Based on the property of combination number, namely, $C_{k+1}^i=C_{k}^i+C_{k}^{i-1}$, we can derive the following:
\begin{equation}
\begin{aligned}	 
\sum_{i=0}^{k}C_{k+1}^i \rho_1^i |\mathbb{S}|^{k-i}  & = 
\sum_{i=0}^{k}(C_{k}^i+C_{k}^{i-1}) \rho_1^i |\mathbb{S}|^{k-i} \\ & =
\sum_{i=0}^{k}C_{k}^i \rho_1^i |\mathbb{S}|^{k-i}+
\sum_{t=1}^{k}C_{k}^{t-1} \rho_1^t |\mathbb{S}|^{k-t} \\ & =
\sum_{i=0}^{k-1}C_{k}^i \rho_1^i |\mathbb{S}|^{k-i}+
\rho_1^k+
\sum_{i=0}^{k-1}C_{k}^{i} \rho_1^{i+1} |\mathbb{S}|^{k-i-1}. \notag
\end{aligned}
\end{equation}

Therefore, we can draw the conclusion that the Formula~(\ref{theorem_b}) is correct for $K=k+1$.
The Formula~(\ref{theorem_b}) is therefore true for every natural number $K$.

\paragraph{Spectral analysis}
\label{sec: sa}
In addition to heuristically understanding the global interactions as feature clusters in the embedding space, we present a more rigorous understanding from the spectral view. As to the artificial graph, the normalized graph Laplacian $\mathbf{L} =
\mathbf{O} - \mathbf{\widetilde{P}} = \mathbf{O}-{\mathbf{Q}^{-\frac{1}{2}}}(\mathbf{P}+\rho \mathbf{O}){\mathbf{Q}^{-\frac{1}{2}}}$.
$\mathbf{L}$ is a symmetric positive semi-definite matrix and can be decomposed into the form of $\mathbf{L}=\mathbf{U}\mathbf{\Lambda}\mathbf{U}^\top$, where $\mathbf{U} \in \mathbb{R}^{|\mathbb{S}| \times |\mathbb{S}|}$ is the matrix composed of orthogonal eigenvectors and $\mathbf{\Lambda}=diag(\lambda_i, 1 \leq i\leq |\mathbb{S}|) \in \mathbb{R}^{|\mathbb{S}| \times |\mathbb{S}|}$ is a diagonal matrix of its eigenvalues.
The graph convolution is equal to:
\begin{equation}\notag
\mathbf{g}* \mathbf{s} = 
\mathbf{U}((\mathbf{U}^\top\mathbf{g})\odot(\mathbf{U}^\top\mathbf{s})) =
\mathbf{U}\mathbf{\hat G}\mathbf{U}^\top\mathbf{s},
\end{equation}
where $\mathbf{s} \in \mathbb{R}^{|\mathbb{S}|}$ denotes a signal to be transformed (each column of $\mathbf{E}$); $\mathbf{g}$ denotes a filter; and $\mathbf{\hat G}=diag(\hat g(\lambda_i), 1 \leq i\leq |\mathbb{S}|)$ represents the diagonal matrix consisting of the spectral filter coefficients $\hat g(\lambda_i)$.
Functionally, the eigenvalues represent different frequencies; $\mathbf{U}^\top\mathbf{s}$ is the projection (decomposition) of signal $\mathbf{s}$ along the frequencies. Upon the decomposition, the graph convolution filters the signals according to the spectral filter coefficients. 

As to our global interaction modeling,
$\mathbf{\widetilde{P}}\mathbf{E} = 
(\mathbf{O} - \mathbf{L})\mathbf{E} = 
(\mathbf{O} - \mathbf{U}\mathbf{\Lambda}\mathbf{U}^\top)\mathbf{E} =
\mathbf{U}(\mathbf{O}-\mathbf{\Lambda})\mathbf{U}^\top\mathbf{E}$.
Thus, for a specific frequency $\lambda_i$, its spectral filter coefficient $\hat g(\lambda_i)=1-\lambda_i$.
Note that the eigenvalues (filter frequencies) of $\mathbf{L}$ are $\lambda_1=0$ and $\lambda_2=|\mathbb{S}| /(|\mathbb{S}|+ \rho)$ ($|\mathbb{S}|-1$ multiplicities), and the corresponding spectral filter coefficients are 1 and $\rho /(|\mathbb{S}|+ \rho)$, respectively.
Here, the filter frequency of $\lambda_1=0$ preserves the original input information, while $\lambda_2=|\mathbb{S}| /(|\mathbb{S}|+ \rho)$ is adjustable, which can filter out the global interaction signal\footnote{
In order to ensure the consistency of the global detection frequency, we need to fix the size of $|\mathbb{S}|$. This is also a reason for sampling a fixed number of features for each node.}.
That is, we can find the frequency $\lambda_2$ where the global interaction signal exists by adjusting probe coefficient $\rho$.
Therefore, it is essential to set a proper probe coefficient $\rho$ in the global interaction modeling (see results in Figure~\ref{fig:lb}), which is a hyper-parameter of CatGCN.

\subsubsection{Node representation fusion}
Aiming to thoroughly exploit the benefit from both local feature interactions and global feature interactions, CatGCN fuses $\mathbf{h}_l$ and $\mathbf{h}_g$ into an overall node representation $\mathbf{h}$ through an aggregation layer. 
As aforementioned, $\mathbf{h} \in \mathbb{R}^{C}$ is the input of the pure neighborhood aggregation and required to be in the label space. 
As such, the aggregation layer is also responsible for projecting the representation into label space. 
Here we perform a late fusion strategy, which adds the two interacted representations after projecting them into the label space: 
\begin{equation}\notag
\mathbf{h}'_l=\sigma(\mathbf{W}_l\mathbf{h}_l+\mathbf{b}_l),
\end{equation}
\begin{equation}\notag
\mathbf{h}'_g=\sigma(\mathbf{W}_g\mathbf{h}_g+\mathbf{b}_g),
\end{equation}
\begin{equation}
\mathbf{h} = \alpha\mathbf{h}'_g+(1-\alpha)\mathbf{h}'_l,
\end{equation}
where $\alpha \in [0,1]$ is a hyper-parameter to balance the influence of local and global interaction modeling, $\mathbf{W}_g$ and $\mathbf{W}_l$ are projection matrices.
Note that we can take multiple fully connected layers here to enhance the expressiveness of the projection while ensuring the last one's output dimension is consistent with the predicted classes.

\subsection{Discussion}
\textit{Relation with Fi-GNN.}
Fi-GNN~\cite{Fi-GNN} is a click-through rate~(CTR) prediction framework adopting GNN module, which also models the global addition-based interaction on an artificial feature graph. 
Fi-GNN adopts graph attention to model the structure of the feature graph, which dynamically calculates the strength of connections for each edge in the graph. 
However, as pointed out in~\cite{knyazev2019understanding}, graph attention is not suitable for this situation which lacks supervised training on attention weights and is hard to find optimal initialization, leading to inferior performance.
Moreover, Fi-GNN further stacks edge information transmission mechanism and recurrent embedding updating mechanism, which poses great challenge on model training, e.g., unaffordable computation and memory cost and severe overfitting. 
By contrast, CatGCN models global feature interactions in a very concise manner, which has the same complexity as a standard fully-connected layer (see Table~\ref{tab:fc} for an in-depth comparison).

\noindent \textit{Relation with APPNP.} 
To best of our knowledge, APPNP~\cite{APPNP} is the first method that decouples the feature transformation and neighborhood aggregation in GCN layers.
The target of APPNP is to alleviate the over-smoothing issue of deep GCN models which can lose focus at the upper layers.
Instead of resolving over-smoothing, CatGCN focuses on enhancing the initial node representation which is a dual perspective. 
More specifically, CatGCN enhances the node representation through integrating two kinds of explicit interactions between categorical features, which has not been studied before.
Further experimental results show that if APPNP unitizes our scheme to obtain the initial node representation, it will bring significant performance improvements (see Figure~\ref{fig:ft} for details).

\section{Experiments}
\label{sec:experiments}

\begin{table}[t]
\centering
\caption{Statistics of the datasets.}
\vspace{-0.3cm}
\resizebox{0.97\columnwidth}{!}{
\begin{tabular}{c|c|c|r|r|r}
\hline 
\multicolumn{1}{c|}{{ \textbf{Dataset}}} & {\textbf{Attribute}} & \multicolumn{1}{c|}{{\textbf{Class}}} & \multicolumn{1}{c|}{{\textbf{Node}}} & \multicolumn{1}{c|}{{\textbf{Feature}}} & \multicolumn{1}{c}{{\textbf{Edge}}} \\
\hline \hline

\multicolumn{1}{c|}{\textbf{Tencent}} & age & 7 & 51,378 & 309 & 64,514 \\ \hline

\multirow{2}{*}{\textbf{Alibaba}}     & purchase & 3 & \multirow{2}{*}{166,958} & \multirow{2}{*}{2,820} & \multirow{2}{*}{14,614,182} \\ \cline{2-3} & city  & 4  &  &  & \\ \hline 
\end{tabular}
}
\vspace{-0.3cm}
\label{tab:sd}
\end{table}

In this section, we conduct extensive empirical studies to investigate the following research questions:
\begin{description}
	\item[RQ1] How does our proposed feature interaction modeling strategies affecting the initial node representation?
	\item[RQ2] How does our CatGCN perform compared to the state-of-the-art methods?
    \item[RQ3] What are the factors that influence the effectiveness of CatGCN?
\end{description}

\subsection{Experimental settings}

\subsubsection{Datasets}

In order to investigate the actual performance of the model, we select three large-scale node classification datasets from real scenes: \textbf{Tencent-age}, \textbf{Alibaba-purchase}, and \textbf{Alibaba-city}~\cite{DIN}. \textbf{Tencent-age} is a social network graph with the target of predicting the user's age level.
\textbf{Alibaba-purchase} and \textbf{Alibaba-city}~\cite{DIN} are also user profiling tasks on an e-commerce platform user graph, where the consumption level and city level are the prediction labels, respectively.
These three datasets are collected from social platform Tencent and e-commerce platform Alibaba, and their construction process is as follows:

\begin{itemize}[leftmargin=*]
\item \textbf{Tencent\footnote{\url{https://www.kaggle.com/c/kddcup2012-track1}}}:
This dataset is provided by the social networking platform Tencent Weibo, which includes users' preferences for a variety of items~(\eg celebrities, organizations, and groups).
We choose these items as the categorical features of user nodes.
A preliminary cleanup of the data (e.g., filtering out users with inappropriate age settings) comes up with 1,238,563 users, which is termed as \textbf{Tencent-large}.
In addition, considering that most GCNs are not designed for handling such large-scale graph, we select a sub-graph with 51,378 nodes of active users with at least 10 interaction on items.
In our processing, if one user have followed the item $i$, we set $x_i=1$, otherwise $x_i=0$.
In this way, we can obtain the multi-hot categorical features $\mathbf{x} \in \mathbb{R}^{d}$ of this user node, where $d=309$ in this dataset.
Although this dataset is provided for the recommendation task, it also provides information about the user age attribute, from which we selected over fifty thousands of users to perform the user profiling node classification task.
Meanwhile, users of social platforms will interact with others in a series of ways, such as thumb up, comment, and forwarding, which leads to straightforward interconnections between users.
In our experiment, we use the "follow" relationship to establish edges between user nodes.
Note that the difference between the followed and following are ignored in our processing, that is, the edges we create are undirected.

\item \textbf{Alibaba\footnote{\url{https://tianchi.aliyun.com/dataset/dataDetail?dataId=56}}}:
This is a dataset of click-through rates for display ads on Alibaba's Taobao platform. 
In this scenario, we choose the categories of products as the categorical features affiliated to user nodes.
Particularly, if one user has clicked products belonging to the category $i$, we set $x_i=1$, otherwise $x_i=0$.
Thus, we acquire the user categorical features $\mathbf{x} \in \mathbb{R}^{d}$ with dimensions $d=2,820$ in Alibaba dataset.
For our user profiling task, we screen two  high-value user attributes, namely purchase, and city, corresponding to consumption level and city level where the user lives.
Since there is no correlation like "follow" between users in the e-commerce platform, we establish the relationship between users based on co-click. 
In other words, if users jointly click the same product, we establish an edge between the two user nodes.
Naturally, the edges between users established through this common behavior are undirected.
\end{itemize}

Statistics for above datasets are shown in Table~\ref{tab:sd}.
In addition, we construct a synthetic graph to investigate the effects of feature interaction modeling on the initial node representation.
We first generate 1,000 nodes and randomly divide them into two classes with equal probability. 
We then sample edges between nodes of the same class from a Bernoulli distribution with a probability of 0.005, and decrease the probability to 0.001 for sampling edges across different classes. In this way, we acquire a synthetic graph with 1,000 nodes and 5,042 edges.
Lastly, we generate a binary feature vector $\mathbf{x}$ with dimension of 100 for each node, where 10 entries are non-zero. 
In particular, we randomly assign non-zero entries within dimension 1-70 for node belongs to the first class, and dimension 31-100 for node in the second class. 
In this way, each class has 30 class-specific features and 40 intermingled features that are clues and barriers to distinguish the classes, respectively. Note that both the local interaction between a class-specific feature and an intermingle feature, and the global interaction can facilitate the classification.
For each dataset, we randomly select 80\%, 10\%, and 10\% of nodes to form the training, validation, and testing set, respectively.

\subsubsection{Baseline models}
We compare CatGCN with several recent GCN models, including the classical methods GCN~\cite{GCN}, GAT~\cite{GAT}, GraphSAGE~\cite{GraphSAGE} and the latest state-of-the-art models APPNP~\cite{APPNP}, SGC~\cite{SGC}, CrossGCN~\cite{Cross-GCN} and GCNII~\cite{GCNII}.
\begin{itemize}[leftmargin=*]
    \item \textbf{GCN}~\cite{GCN}: 
    This is a semi-supervised classification model on graph-structured data, which can effectively aggregate information from the neighborhood by simultaneously encoding the graph structure and node features.
    \item \textbf{GAT}~\cite{GAT}:
    It adaptively allocates weight to neighborhood nodes through the masked self-attention layer, so as to distinguish the importance of different neighborhood nodes when aggregating neighborhood information.
    \item \textbf{GraphSAGE}~\cite{GraphSAGE}: 
    This method implements representation learning on the large-scale graph by sampling local neighborhoods of nodes.
    In our experiment, we use the mean aggregator to complete the aggregation of neighborhood information.
    \item \textbf{APPNP}~\cite{APPNP}: 
     It connects GCN with Personalized PageRank~\cite{haveliwala2002topic} and expands the available neighborhood range without introducing additional parameters.
    \item \textbf{SGC}~\cite{SGC}: 
    This method eliminates unnecessary nonlinearities and weight matrices in GCN and effectively reduces the complexity of the model.
    It not only improves performance but also significantly reduces computing costs.
    \item \textbf{CrossGCN}~\cite{Cross-GCN}: 
    This method obtains cross features based on the traditional matrix factorization approach to enhance the feature learning ability.
    However, it can only model local feature interactions.
    \item \textbf{GCNII}~\cite{GCNII}: 
    It introduces initial residual connection and identity mapping to prevent the over-smoothing problem of GCN, which enables the model to be deeply stacked and brings performance gains.
\end{itemize}

For all aforementioned models, we re-implement them using PyTorch Geometric~\cite{PyG}, which has consistent or even better performance than the original paper.
Our implementations are available at https://github.com/TachiChan/CatGCN.

\subsubsection{Parameter settings}
For all methods, the dimension of the categorical feature embedding layer and the size of all hidden layers are set to 64 for a fair comparison.
All trainable parameters are initialized with the Xavier method~\cite{Xavier} and optimized with Adam~\cite{Adam}.
We apply grid search strategy for hyper-parameters: the learning rate is tuned among $\{1e{-1}, 1e{-2}, 1e{-3}\}$, the $L_2$ regularization coefficient is searched in the range of $\{1e{-5}, 1e{-4}, ... , 1e{-1}, 0.0\}$, and dropout ratio is tuned in $\{0.0, 0.1, ..., 0.9\}$. 
For all baseline methods, their node representations are aggregated from the node's associated categorical features in the way of mean pooling, which is the normalized version of aforementioned $\textbf{H}^{(0)} \textbf{W}^{(0)}$.
For CatGCN, we take ReLU as the activation function $\sigma$, and tune the aggregation parameter $\alpha$ within $\{0.0, 0.1, ..., 0.9, 1.0\}$.
For each node, we sample a fixed number of categorical features from $\mathbb{S}$ to maintain a consistent global probe frequency $\lambda_2$, which is set to 10 in our experiment.
The optimal hyper-parameters of CatGCN on different datasets are listed in Table~\ref{tab:hp}.
In all cases we adopt an early stopping strategy on the validation set with a patience of 10 epochs, and report the testing \textit{Accuracy} and \textit{Macro-F$_1$}~\cite{HGAT,wu2019neural}.

\begin{table}[t]
\centering
\caption{
The optimal hyper-parameters of CatGCN on different datasets where lr and dr are short for learning rate and dropout rate, respectively.
}
\vspace{-0.3cm}
\resizebox{0.85\columnwidth}{!}{
\begin{tabular}{c r r r r r r}
\hline
\multicolumn{1}{c}{\textbf{Dataset}} & 
\multicolumn{1}{c}{\textbf{lr}} & 
\multicolumn{1}{c}{\textbf{$L_2$}} & 
\multicolumn{1}{c}{\textbf{dr}} & 
\multicolumn{1}{c}{\textbf{$\rho$}} & 
\multicolumn{1}{c}{\textbf{$\alpha$}} & 
\multicolumn{1}{c}{\textbf{$L$}} \\ \hline
\textbf{Tencent-age}     & $0.1$ & $1e-4$ & 0.3 & 1 & 0.4 & 6 \\ 
\textbf{Alibaba-purchase} & $0.1$ & $1e-5$ & 0.3 & 39 & 0.9 & 8 \\
\textbf{Alibaba-city}     & $0.1$ & $1e-5$ & 0.9 & 41 & 0.3 & 3 \\ \hline
\end{tabular}}
\vspace{-0.3cm}
\label{tab:hp}
\end{table}

\subsection{Effects on node representation quality (RQ1)}
\label{sec:rq1}

\begin{figure}[]
	\centering
	\mbox{
		\hspace{-0.15in}
		\subfigure[Silhouette Score]{
			\label{fig:hs}
			\includegraphics[width=0.26\textwidth]{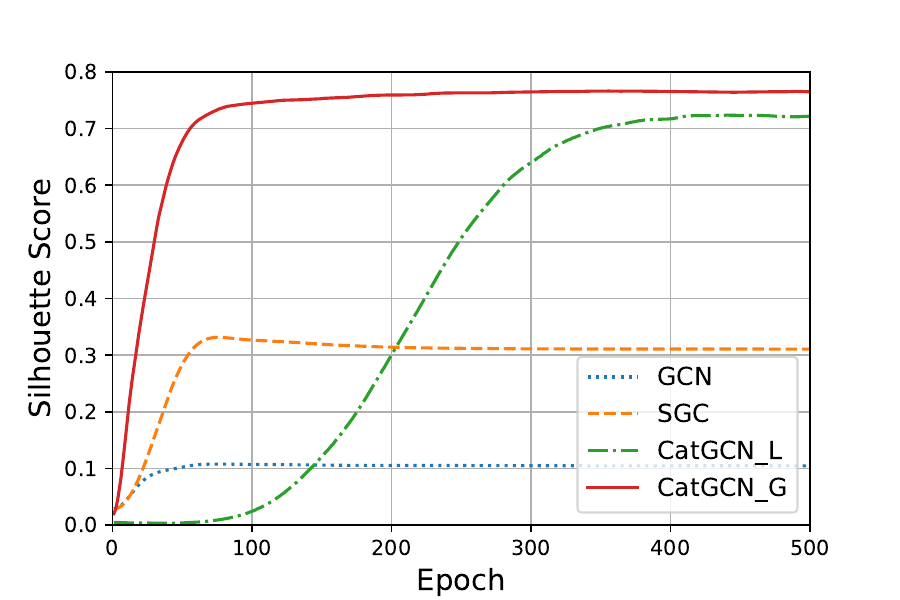}
		}
		\hspace{-0.21in}
		\subfigure[t-SNE visualization]{
			\label{fig:tsne}
			\includegraphics[width=0.26\textwidth]{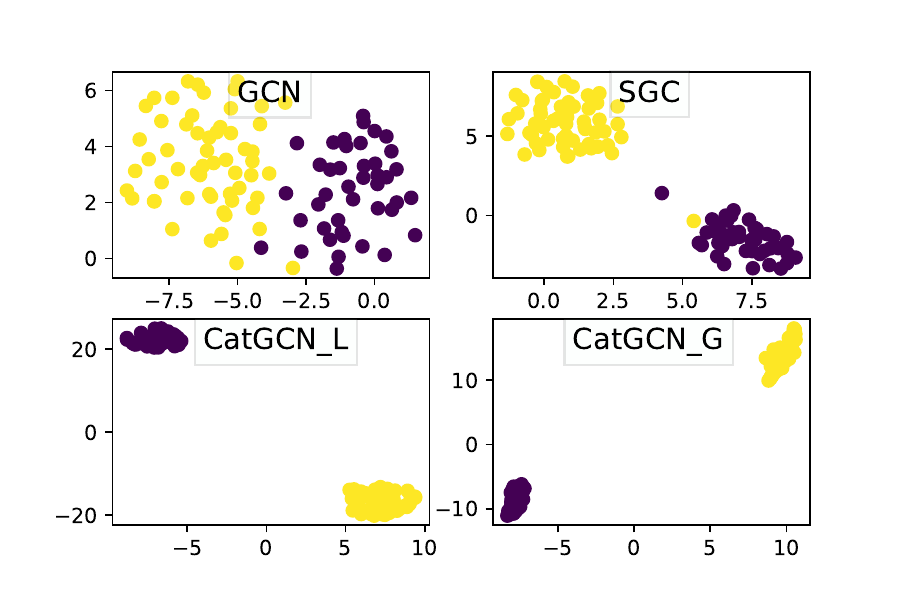}
		}
	}
	\vspace{-0.3cm}
	\caption{(a) The Silhouette Score of initial node representation on the synthetic dataset during the training process. (b) The t-SNE visualization of the initial node representation at the end of training, i.e., Epoch 500. 
	}
	\label{fig:repr}
	\vspace{-0.4cm}
\end{figure}

To investigate the influence of feature interaction modeling on the initial node representation, we test the following four models on the synthetic dataset: GCN~\cite{GCN}, SGC~\cite{SGC}, CatGCN\_L and CatGCN\_G.
CatGCN\_L and CatGCN\_G are variants of CatGCN which calculate the initial node representation with local and global feature interaction modeling, respectively. As to GCN and SGC, they obtain the initial node representation from the normal feature transformation without explicit feature interaction modeling.
Note that we set the same dimension for the initial node representation of the four models.
In particular, all models are trained with 500 epochs where we extract the initial node representations of the testing nodes at each epoch.

\begin{table*}[t]
\centering
\caption{Node classification performance of all compared methods on the three real-world datasets.}
\vspace{-0.3cm}
\resizebox{1.97\columnwidth}{!}{
\begin{tabular}{c|c|c|c|c|c|c c}
\hline \hline
 \multicolumn{1}{c|}{\textbf{Dataset}} & \multicolumn{2}{c|}{\textbf{Tencent-age}} & 
 \multicolumn{2}{c|}{\textbf{Alibaba-purchase}} &
 \multicolumn{2}{c}{\textbf{Alibaba-city}} &\\ \hline
 \textbf{Method} & 
 \textbf{Accuracy} & \textbf{Macro-F$_1$} &  \textbf{Accuracy} & \textbf{Macro-F$_1$} &
 \textbf{Accuracy} & \textbf{Macro-F$_1$} \\ \hline\hline
 
 \textbf{GCN} & $0.2014  \textit{(+24.6\%)}$ & $0.1586 \textit{(+20.2\%)}$ & $0.4420 \textit{(+25.9\%)}$ & $0.3904 \textit{(+14.9\%)}$ & $0.2648 \textit{(+30.6\%)}$ & $0.2585 \textit{(+9.2\%)}$\\ \hline
 
 \textbf{GAT} & $0.2347 \textit{(+\ 6.9\%)}$ & $0.1740 \textit{(+\ 9.5\%)}$ & $0.4677 \textit{(+19.0\%)}$ & $0.4238  \textit{(+\ 5.8\%)}$ & $0.3313 \textit{(+\ 4.4\%)}$ & $0.2779 \textit{(+1.5\%)}$\\ \hline
 
 \textbf{GraphSAGE} & $0.2386 \textit{(+\ 5.2\%)}$ & $0.1769 \textit{(+\ 7.7\%)}$ & $0.4863 \textit{(+14.4\%)}$ & $0.4174 \textit{(+\ 7.4\%)}$ & $0.2895 \textit{(+19.4\%)}$ & $0.2719 \textit{(+3.8\%)}$\\ \hline
 
 \textbf{APPNP} & $0.2472 \textit{(+\ 1.5\%)}$ & $0.1822 \textit{(+\ 4.4\%)}$ & $0.4860 \textit{(+14.5\%)}$ & $0.3939 \textit{(+13.8\%)}$ & $0.3066 \textit{(+12.8\%)}$ & $0.2692 \textit{(+4.8\%)}$ \\ \hline
 
 \textbf{SGC} & $0.2411 \textit{(+\ 4.1\%)}$ & $0.1777 \textit{(+\ 7.3\%)}$ & $0.4832 \textit{(+15.1\%)}$ & $0.4167 \textit{(+\ 7.6\%)}$ &  $0.2880 \textit{(+20.1\%)}$ & $0.2717 \textit{(+3.9\%)}$\\ \hline
 
 \textbf{CrossGCN} & $0.2238 \textit{(+12.1\%)}$ & $0.1721 \textit{(+10.7\%)}$ & $0.3980 \textit{(+39.8\%)}$ &  $0.3593 \textit{(+24.8\%)}$ & $0.3114 \textit{(+11.0\%)}$ & $0.2776 \textit{(+1.7\%)}$\\ \hline 
 
 \textbf{GCNII} & $0.2310 \textit{(+\ 7.9\%)}$ & $0.1777 \textit{(+\ 7.3\%)}$ & $0.4275 \textit{(+23.2\%)}$ &  $0.3778 \textit{(+18.6\%)}$ & $0.2925 \textit{(+18.2\%)}$ & $0.2669 \textit{(+5.7\%)}$\\ \hline 
 
 \textbf{CatGCN(ours)} & $\textbf{0.2509}$ & $\textbf{0.1906}$ & $\textbf{0.5564}$ &  $\textbf{0.4484}$ & $\textbf{0.3458}$ & $\textbf{0.2822}$ \\ \hline \hline
\end{tabular}}
\vspace{-0.3cm}
\label{tab:pc}
\end{table*}

Quantitatively, we evaluate the node representation quality through the Silhouette Score~\cite{rouss1987silho}, which is defined over the set of testing nodes: 
\begin{equation}
s = mean\left(\Big\{
    \frac{b(u)-a(u)}{max(a(u),b(u))}
    \Big\}\right),
\end{equation}
where $a(u)$ is the mean intra-class distance of node $u$, i.e., the average distance between node $u$ and the other nodes in the same class as $u$; and $b(u)$ is the mean inter-class distance of node $u$. Note that the value of Silhouette Score is in the range of [-1,1] where a larger value means the nodes in different classes are separated, i.e., the better node representation.
Figure \ref{fig:hs} shows the Silhouette Score of the four testing models along their training procedure. From the figure, we can see that both CatGCN\_G and CatGCN\_L can achieve higher Silhouette Score compared to GCN and SGC, which shows the benefit of feature interaction modeling. 
Moreover, we qualitatively evaluate the initial node representation by performing dimension reduction through t-SNE~\cite{tSNE} and visualizing the nodes.
Figure \ref{fig:tsne} illustrates the node representation of the testing models at epoch 500.
From the figure, we can see that the initial node representation affiliated to different classes of CatGCN\_L and CatGCN\_G has a significantly higher distinction than GCN and SGC, which further reflects the strength of our proposed local and global feature interaction strategies.

\subsection{Overall performance comparison (RQ2)}
Table~\ref{tab:pc} shows the testing performance of all compared methods on the three datasets.
From the table, we have the following observations:
\begin{itemize}[leftmargin=*]
 \item In all cases, CatGCN outperforms all baselines with a significant gain of 12.41\% on average, which is attributed to incorporating both the local and global feature interactions into the initial node representations. 
 As such, this result validates the rationality of explicit interaction modeling of categorical features in GCN models. 
 \item GAT performs better than the standard GCN, which shows the benefit of graph attention in these tasks. 
 CatGCN may also achieve better performance if using attention in its PNA module, which is discarded purely for the consideration of computation cost.
 \item Many SOTA models fail to achieve ideal performance, and none of them can deliver consistently superior performance across all tasks.
 Note that in Alibaba-city task, GAT exceeds all the benchmark schemes, indicating that the adjacent nodes in the dataset may not meet the similarity, which is also a common phenomenon in real scenes.
 Existing models are designed from the perspective of neighborhood aggregation on the graph, so it is difficult to maintain stable performance in complex scenarios.
 \item Our feature interaction modeling is equivalent to adding two types of valuable input information to the initial node representation, one is the combination features, and the other is the global peculiarity information.
 This approach can increase the distinction of node representation and thus alleviate the interference caused by the noisy edge.
 The corresponding experimental results strongly support this claim (CatGCN consistently exceeds all baselines).
\end{itemize}

\textbf{Large-scale graph.} Recall that our target is to obtain a better initial node representation, and the feature interaction modeling strategies can also be applied to GCNs designed for handling large-scale graphs. 
In this light, we combine feature interaction modeling part into Cluster-GCN~\cite{ClusterGCN} and compare it with the vanilla Cluster-GCN on \textbf{Tencent-large}.
After such operation, the \textit{Accuracy} is improved from 0.2321 to 0.2419, and the \textit{Macro-F$_1$} is improved from 0.176 to 0.177.
This further verifies the applicability and effectiveness of our proposed framework in real massive data scenarios.

\begin{figure*}[t]
	\centering
	\subfigure{\includegraphics[width=0.3\textwidth]{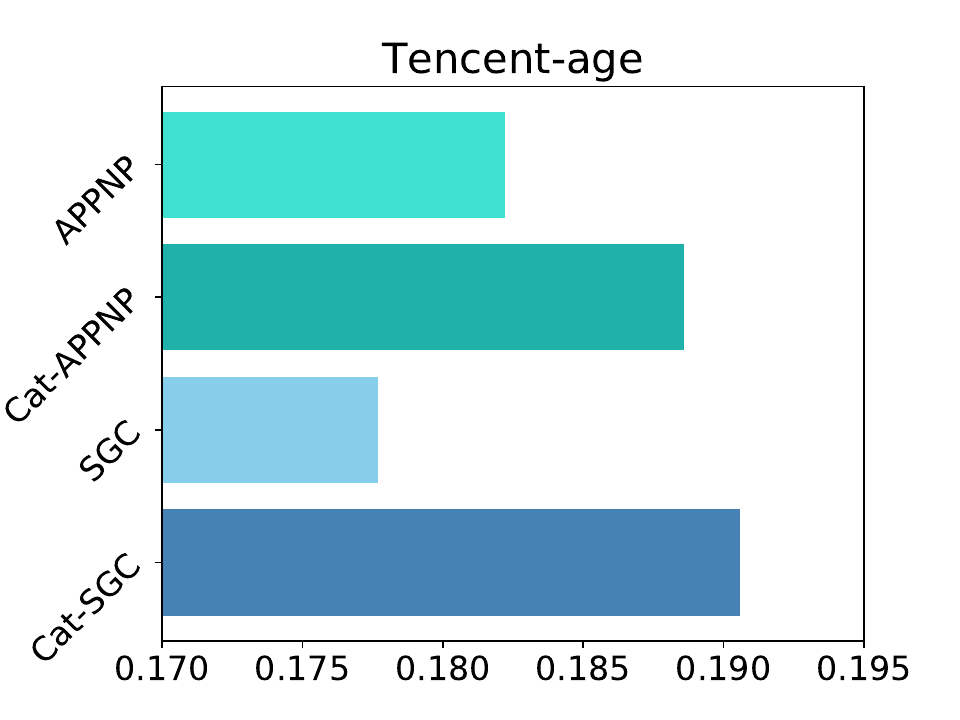}} 
	\subfigure{\includegraphics[width=0.3\textwidth]{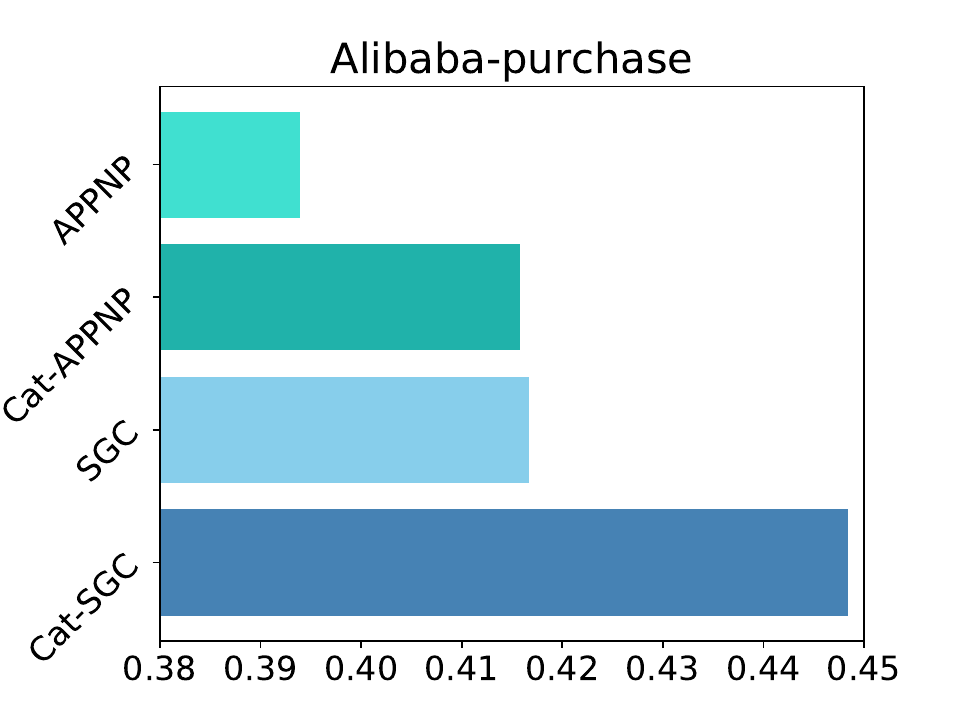}} 
	\subfigure{\includegraphics[width=0.3\textwidth]{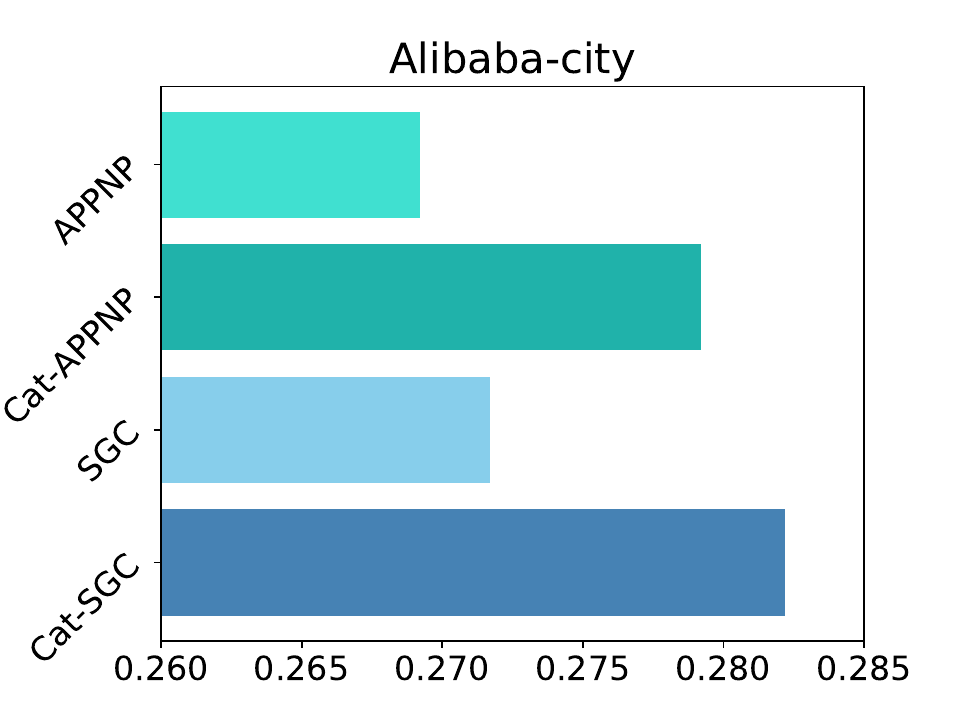}}
	\vspace{-10pt}
	\caption{Impacts of feature interaction modeling.}
 	\vspace{-10pt}
	\label{fig:ft}
\end{figure*}

\begin{figure*}[t]
	\centering
	\subfigure{\includegraphics[width=0.3\textwidth]{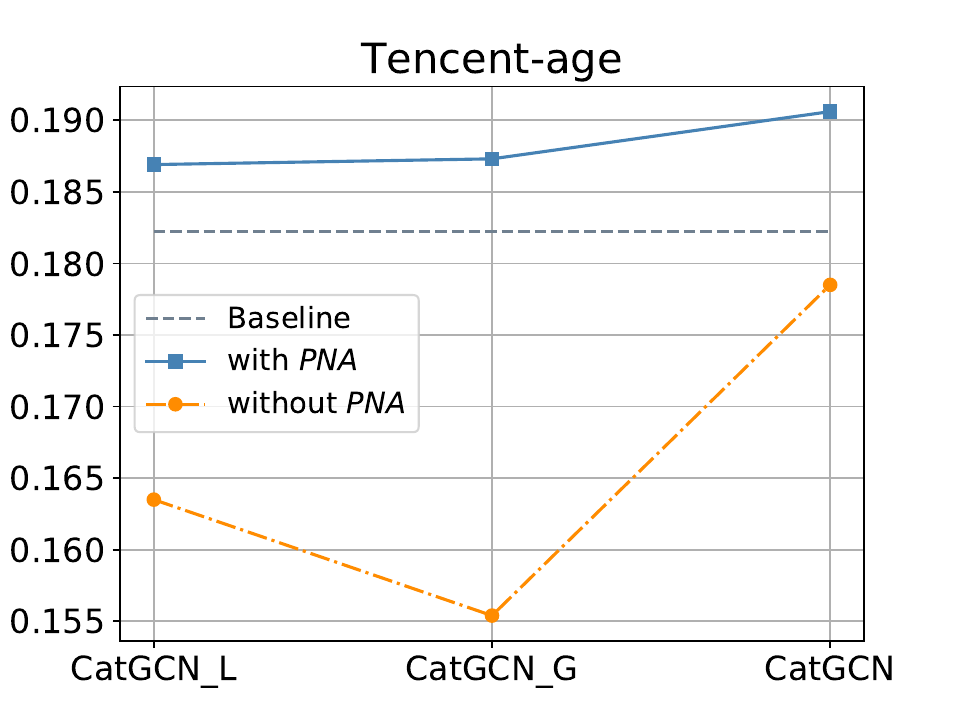}} 
	\subfigure{\includegraphics[width=0.3\textwidth]{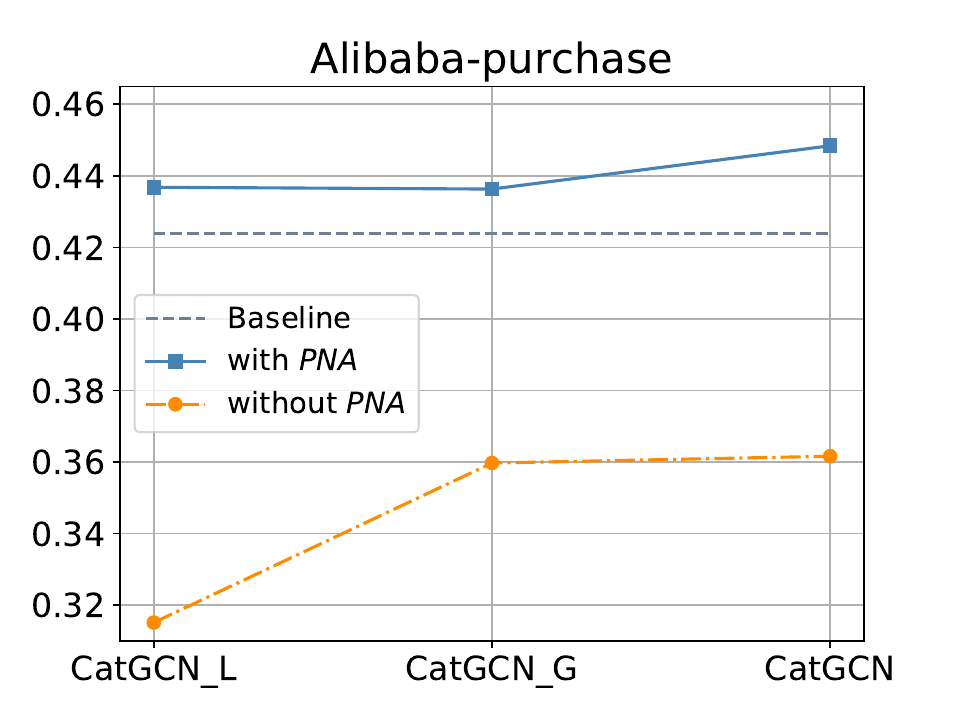}} 
	\subfigure{\includegraphics[width=0.3\textwidth]{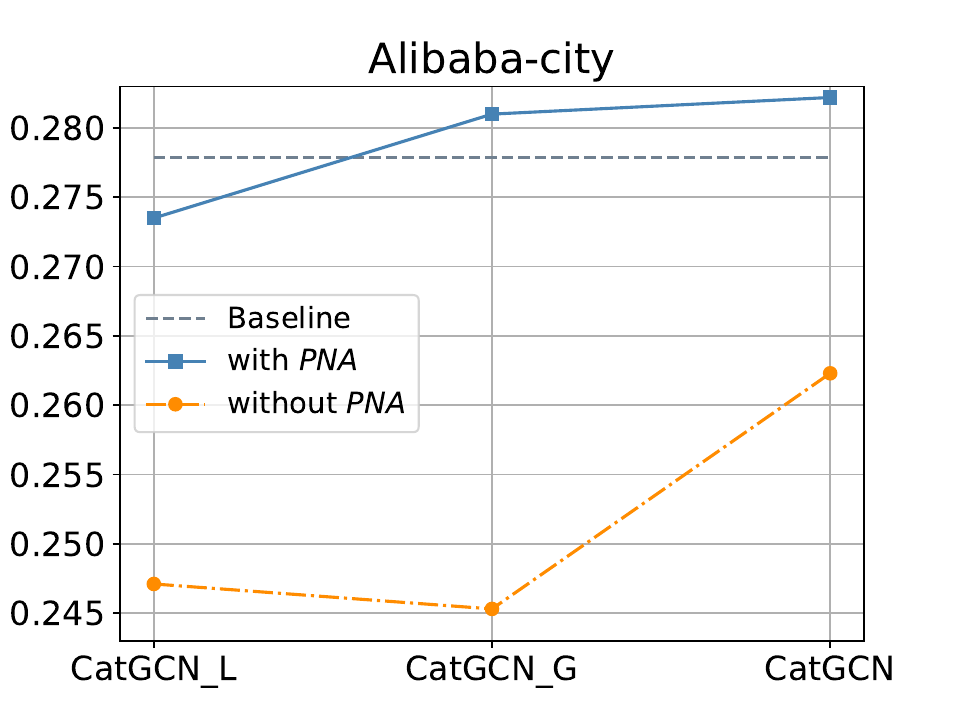}}
	\vspace{-10pt}
	\caption{Impacts of pure neighborhood aggregation.} 
	\label{fig:na}
 	\vspace{-10pt}
\end{figure*}

\subsection{In-depth analysis (RQ3)}
To further validate the rationality of our model design, we separately test the feature interaction modeling modules and the pure neighborhood aggregation module.
To save space, we omit the results \wrt \textit{Accuracy}, which have the similar trend as \textit{Macro-F$_1$}.

\subsubsection{Impacts of feature interaction modeling}
We equip APPNP with the feature interaction modeling part of CatGCN~(i.e., the local and global interaction modeling), which is named as Cat-APPNP. 
Figure~\ref{fig:ft} shows the performance of standard APPNP, Cat-APPNP, SGC, and CatGCN (i.e., Cat-SGC). 
Note that SGC is equivalent to CatGCN without feature interaction modeling.
As can be seen, Cat-APPNP and Cat-SGC significantly outperform the corresponding APPNP and SGC, which further validates the effectiveness of enhancing initial node representation in GCN models and the advantages of modeling categorical features interactions.
Moreover, the improvement of Cat-SGC over SGC is larger than that of Cat-APPNP over APPNP in all cases.
We postulate the reason is that SGC and APPNP actually adopt different feature transformation strategies and the MLP adopted by APPNP can account for some feature interactions implicitly.
Accordingly, applying the feature interaction modeling module can bring greater improvement to the initial node representation of SGC.

Furthermore, we develop two variants of CatGCN by removing the global and local interaction modeling mechanisms, which are named CatGCN\_L and CatGCN\_G, respectively.
Figure~\ref{fig:na} shows the performance of CatGCN\_L, CatGCN\_G, and CatGCN (see blue line), where the best result performance across all baselines is also depicted for better comparison (see grey line).
It can be seen that removing any interaction modeling module from CatGCN will lead to performance degradation.
At the same time, both variants outperform or rival all benchmark models. 
Therefore, both local and global interaction modeling mechanisms are effective for node representation learning, and their roles may be complementary.
In addition, we can see that different variants competing on different tasks, which may be related to the different importance of local and global interaction information for different tasks.
Note that if we look only at the feature interaction modeling part, our design can be understood as a plug-and-play framework that can be seamlessly integrated with the existing GNN models(e.g., Cat-APPNP).

\subsubsection{Impacts of pure neighborhood aggregation}
In order to analyze the role of neighborhood aggregation, we remove the PNA module of CatGCN and its two variants CatGCN\_L and CatGCN\_G. 
The corresponding result is shown in Figure~\ref{fig:na}.
The dotted line represents the variation without using the pure neighborhood aggregation~(i.e., without PNA).
The comparison results show that neighborhood aggregation can effectively utilize the network structure to optimize the node representations even if there are no training parameters available, which illustrates that it can not be ignored in the graph convolution models.

\subsubsection{Analysis on the global interaction modeling}
To justify the advantages of the proposed global interaction modeling, we further study the impacts of probe coefficient ($\rho$) and perform in-depth comparison among CatGCN, Fi-GNN, and other global interaction modeling methods.

\begin{figure}[t]
 \centering
 \includegraphics[width=0.7\columnwidth]{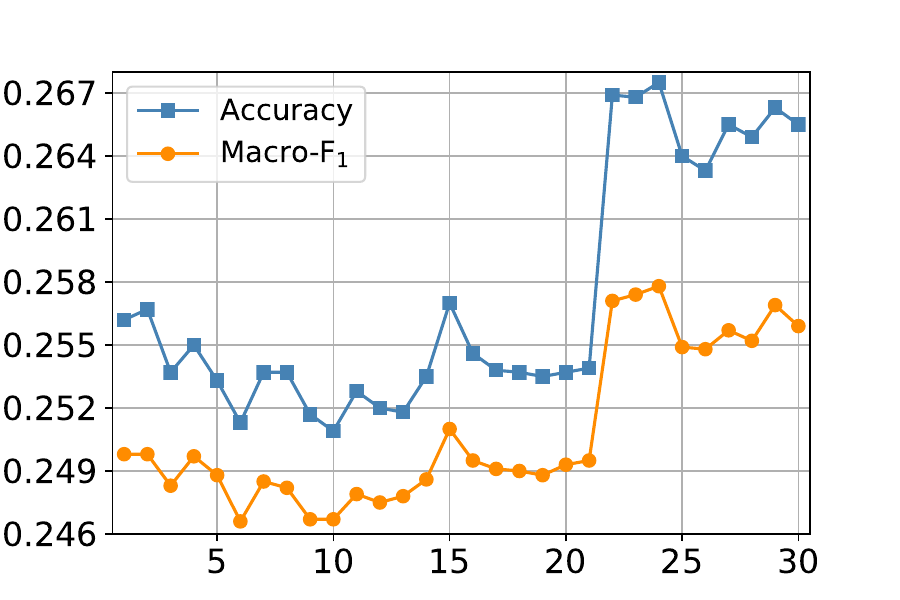}
  \vspace{-10pt}
 \caption{Performance of CatGCN\_G ($L$=1) as adjusting the probe coefficient $\rho$ on Alibaba-city task.
 }
 \vspace{-10pt}
 \label{fig:lb}
\end{figure}

\paragraph{Impacts of probe coefficient ($\rho$)}
We first study how the probe coefficient influences the effectiveness of the proposed global feature interaction modeling. 
Figure~\ref{fig:lb} shows the performance of CatGCN\_G as adjusting the probe coefficient $\rho$ from 1 to 30. 
Note that we test CatGCN\_G so as to avoid the interference of local interaction modeling. Moreover, CatGCN\_G is set to $L$=1 without stacking multiple fully connected layers. 
From the figure, we can observe that: 
1) the performance of CatGCN\_G varies in a large range (0.246\--0.268), which indicates the importance of integrating global peculiarity signal;
2) when $\rho$ exceeds 21, the increase is relatively obvious, indicating that the frequency of global peculiarity signal $\lambda_2$ is around here under current settings;
3) the performance remains at a high level at $\rho \in [22,30]$, possibly because the variation of $\lambda_2$ in this numerical interval is very limited ($\lambda_2=|\mathbb{S}| /(|\mathbb{S}|+ \rho)$).

\paragraph{Comparisons with Fi-GNN}
To demonstrate the superiority of our design in modeling the global feature interactions, we further compare CatGCN with Fi-GNN w.r.t. performance, GPU memory usage and average time per epoch.
For fair comparison, we adopt the same PNA module on the node representation outputted by the interaction modeling mechanism of Fi-GNN (named Fi-SGC).
Due to the huge computational overhead of Fi-SGC's feature interaction modeling mechanism, it still cannot directly run on our dataset, which has a large number of categorical features.
To tackle this issue, we let Fi-SGC share the field-specific weight matrix to reduce the memory requirements.
Even so, the Fi-SGC can only be tested on Tencent dataset, while it will run out of memory on Alibaba dataset.
The experiment results of Tencent dataset are shown in Table~\ref{tab:fc}.
From the table, we can find the hand-crafted design of Fi-GNN doesn't obtain higher performance.
The complex design of the Fi-GNN not only consumes a lot of memory and increases computation time, but also results in performance degradation~(compared to SGC).
As a comparison, Cat-SGC~(i.e., CatGCN) requires about half GPU memory usage and time cost, while significantly improving performance.

Furthermore, we test two variants of CatGCN\_G by replacing the global interaction modeling part with similarity matrix and graph attention mechanism~\cite{GAT}, which are termed as SIMI-SGC and GAT-SGC, respectively.
According to the experimental results in Table~\ref{tab:fc}, SIMI-SGC, GAT-SGC and Fi-SGC have similar performance and are all inferior to SGC. 
Considering that they all use the relationship between feature embeddings to complete the optimization, such training strategy without supervision makes them hard to optimize and leads to poor performance~\cite{knyazev2019understanding}.
This result thus indicates the advantage of performing global interaction modeling in the spectral domain.

\section{Related Work}
\label{sec:related}

As this work explores the modeling of feature interactions in GCN, we review the recent researches on GCN and feature interaction modeling.
\subsection{Graph convolutional networks} 
Graph convolutional networks have recently made remarkable achievements in a series of tasks such as node/graph classification~\cite{GCN,GCNII,feng2019graph,zhuang2020smart,liu2021combining}, link prediction~\cite{zhang2018link, huang2021knowledge,xia2021graph,yang2020learning}, and community detection~\cite{chen2019supervised,he2020community}.
Through coupling feature transformation and neighborhood aggregation, node features and graph structures are encoded simultaneously on each graph convolution layer, which ensures their ability to integrate information on the graph.
To further improve the capability of graph convolutional networks, some strategies are proposed, such as introducing attention mechanisms to distinguish the node contribution~\cite{GAT}, performing node sampling to increase the model scalability~\cite{GraphSAGE,FastGCN,gao2018large}, and simplifying the model framework to reduce the computational cost~\cite{SGC}.
Our work continues the idea of APPNP~\cite{APPNP}, which implies the separation of feature transformation and neighborhood aggregation is a better choice.
We have made an in-depth exploration of the categorical node features, and the proposed framework CatGCN can well adapt to such graph data and obtain the most advanced performance.

\subsection{Feature interaction modeling} 
Feature interactions are critical for revealing intrinsic peculiarity of the node that features affiliated, and they have been extensively explored, especially in real-world applications such as recommendation systems.
The local feature interaction can help enrich valid feature information, and its effectiveness has been verified in several works~\cite{FM,FFM,AFM}.
On the other hand, plenty of researches~\cite{PNN,CFM,CrossNet,yang2019interpretable} have illustrated the importance of global feature interaction modeling.
Further studies~\cite{DeepFM,xDeepFM} demonstrate that the combination of different levels of information can result in improved performance.
Recently, with the rise of graph representation learning, the method of using graph neural network to model feature interactions has appeared, which achieves a good performance in the click-through rate prediction task~\cite{Fi-GNN}.
In our work, we design two different mechanisms to learn the above different levels of information for the categorical node features.
Specifically, for local interactions, we absorb the existing mature work bi-interaction pooling~\cite{NFM}, while for global interactions, we design a specific graph convolutional network based on the nature of categorical feature interactions.
The proposed model that combines these two mechanisms achieves optimal performance while remaining lightweight.
Unlike the route described above, network representation methods with node attributes usually learn the node representation by jointly modeling the structure and attribute information of the network~\cite{HetGNN,ANRL,gao2018deep,meng2019co,liao2018attributed,NAIE}.
Despite their encouraging success, little consideration has been given to using feature interaction modeling to optimize node representation.
By contrast, CatGCN completes node representation optimization based on the node-specific feature set, which is orthogonal to the former design.

\begin{table}[t]
\centering
\caption{Comparison of different global interaction modeling strategies on Tencent-age task.}
\vspace{-0.3cm}
\resizebox{0.97\columnwidth}{!}{
\begin{tabular}{c|r|r|r|r}
\hline
\multicolumn{1}{c|}{\textbf{Methods}} & 
\multicolumn{1}{c|}{\textbf{Accuracy}} & 
\multicolumn{1}{c|}{\textbf{Macro-F$_1$}} & 
\multicolumn{1}{c|}{\textbf{Memory usage}} & 
\multicolumn{1}{c}{\textbf{Time cost}} \\ \hline \hline
\textbf{SGC}     & $0.2411$ & $0.1777$ & 989MB  & 0.03s \\ \hline \hline
\textbf{SIMI-SGC}  & $0.2398$ & $0.1764$ & 1595MB & 0.06s \\ \hline
\textbf{GAT-SGC} & $0.2228$ & $0.1737$ & 3901MB & 0.11s \\ \hline
\textbf{Fi-SGC}  & $0.2234$ & $0.1719$ & 4053MB & 0.11s \\ \hline \hline
\textbf{Cat-SGC} & $0.2509$ & $0.1906$ & 2421MB & 0.06s \\ \hline 
\textbf{CatGCN\_G} & $0.2476$ & $0.1873$ & 2295MB & 0.05s \\ \hline 
\end{tabular}}
\scriptsize{*The testing platform is a Nvidia 2080Ti GPU with an Intel Core i9-9900X CPU~(3.70GHz).}
\label{tab:fc}
\end{table}
\section{Conclusions}
\label{sec:conclusions}

For the scenario of graph learning with categorical node features, we propose a novel GCN model named CatGCN.
By designing local and global feature interaction modeling mechanisms explicitly, our proposed model can fully exploit the information of categorical features, and further integrate their advantages through differentiated aggregation, thus achieving significant improvement in multiple tasks on three large public datasets.
Our proposed model has two cat-like strengths: lightweight (our design can achieve excellent performance with few parameters) and flexibility (the feature interaction modeling part can be seamlessly integrated with the existing GNN models to enhance their performance).
Therefore, the proposed model has great potential in various real-world applications.
In the future, we will incorporate more neighborhood aggregation techniques into CatGCN such as the graph attention and edge dropout~\cite{GAT}.
At the same time, we will consider applying CatGCN to more practical applications, such as the recommender system~\cite{NGCF}, which might be an interesting direction.

\appendices

\ifCLASSOPTIONcompsoc
  \section*{Acknowledgments}
\else
  \section*{Acknowledgment}
\fi

This work is supported by the National Key Research and Development Program of China (2020AAA0106000) and the National Natural Science Foundation of China (62121002, U19A2079).

\ifCLASSOPTIONcaptionsoff
  \newpage
\fi



\bibliographystyle{IEEEtran}
\bibliography{reference}
\end{document}